\documentclass{article} 

\usepackage{iclr2024_conference,times}

\usepackage{amsmath,amsfonts,bm}

\def\eqref#1{equation~\ref{#1}}

\def\1{\bm{1}}

\DeclareMathAlphabet{\mathsfit}{\encodingdefault}{\sfdefault}{m}{sl}
\SetMathAlphabet{\mathsfit}{bold}{\encodingdefault}{\sfdefault}{bx}{n}

\usepackage{hyperref}
\usepackage{url}

\usepackage{xspace}
\usepackage{booktabs}
\usepackage{graphicx}
\usepackage{algorithm}
\usepackage{multirow}
\usepackage{algorithmic}
\usepackage{pgfplots}
\usepackage{makecell}
\usepackage{cleveref}
\usepackage{listings}
\usepackage{titlesec}
\usepackage{amsmath}
\usepackage{pifont}
\usepackage{arydshln}
\usepackage{enumitem}

\crefname{section}{§}{§§}
\Crefname{section}{§}{§§}

\title{%
  ProAgent: From Robotic Process Automation to Agentic Process Automation%
}

\author{Yining Ye$^{1}\thanks{\ \ Indicates equal contribution.}$\hspace{0.5em}, Xin Cong$^{1*}\thanks{\ \  Corresponding author.}$\hspace{0.5em}, Shizuo Tian$^1$, Jiannan Cao$^2$, Hao Wang$^3$, Yujia Qin$^1$, \\ 
\textbf{Yaxi Lu}$^1$, \textbf{Heyang Yu}$^1$, \textbf{Huadong Wang}$^4$, \textbf{Yankai Lin$^5$, Zhiyuan Liu$^{1\dag}$, Maosong Sun$^{1}$} \\
$^1$Tsinghua University $^2$Massachusetts Institute of Technology \\
$^3$Carnegie Mellon University $^4$ModelBest Inc. $^5$Renmin University of China \\
\\
\texttt{\small yeyn2001@gmail.com, xin.cong@outlook.com} 
\\
\texttt{\small tsz21@mails.tsinghua.edu.cn, jiannan@mit.edu, hwang4@alumni.cmu.edu}
}

\definecolor{DarkGreen}{RGB}{30,130,30}
\newcommand{\cmark}{\textcolor{DarkGreen}{\ding{51}}}
\newcommand{\xmark}{\textcolor{red}{\ding{55}}}%

\newcommand{\agenticauto}{\textsc{Agentic Process Automation}\xspace}
\newcommand{\model}{\textsc{ProAgent}\xspace}
\newcommand{\workflow}{{Agentic Workflow Description Language}\xspace} 
\newcommand{\dataagent}{{DataAgent}\xspace} 
\newcommand{\controlagent}{{ControlAgent}\xspace} 

\iclrfinalcopy 

\begin{document}

\maketitle

\begin{abstract}

    From ancient water wheels to robotic process automation~(RPA), automation technology has evolved throughout history to liberate human beings from arduous tasks. 
    Yet, RPA struggles with tasks needing human-like intelligence, especially in elaborate design of workflow construction and dynamic decision-making in workflow execution. 
    As Large Language Models~(LLMs) have emerged human-like intelligence, this paper introduces \agenticauto(APA), a groundbreaking automation paradigm using LLM-based agents for advanced automation by offloading the human labor to agents associated with construction and execution. 
    We then instantiate \model, an LLM-based agent designed to craft workflows from human instructions and make intricate decisions by coordinating specialized agents. 
    %
    %
    Empirical experiments are conducted to detail its construction and execution procedure of workflow, showcasing the feasibility of APA, unveiling the possibility of a new paradigm of automation driven by agents.
    Our code is public at \href{https://github.com/OpenBMB/ProAgent}{https://github.com/OpenBMB/ProAgent}.

\end{abstract}

\begin{figure}[!h]
    \centering
    \vspace{-0.6cm}
    \includegraphics[width=1.0\linewidth]{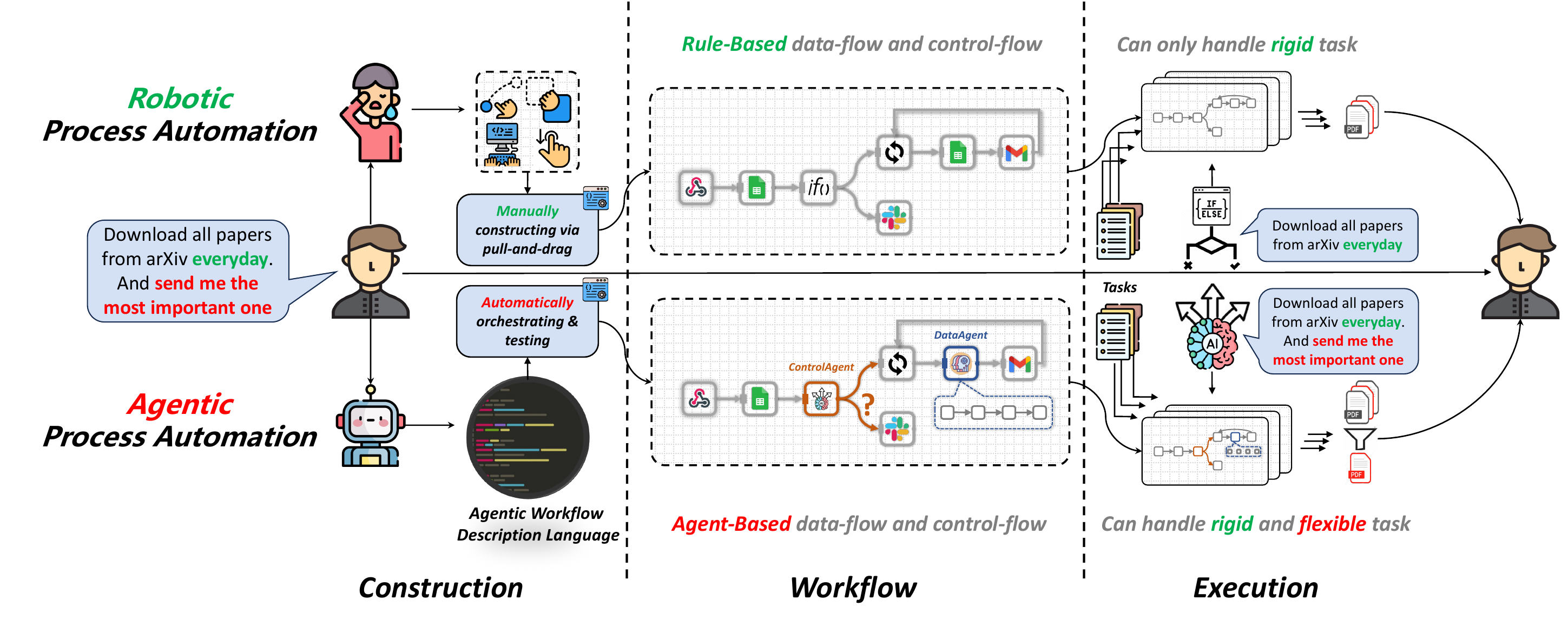}
    \vspace{-0.35cm}
    \caption{The comparison between Robotic Process Automation and Agentic Process Automation.}
    \label{fig:framework-comparison}
\end{figure}

\section{Introduction}

Automation, aiming to reduce human intervention in processes and enhance efficiency, has undergone a series of evolutionary stages throughout history.
From the waterwheel irrigation system in the early agricultural age to steam engines in the industrial age, the human race has continuously been pursuing to offload human labor to autonomous systems, liberating themselves from arduous processes.
Entering the information age, marked by a rapid shift from traditional industry to an economy primarily based on digital technology, software has been widely used as it serves as the foundation for the processing, storage, and communication of information.
Robotic Process Automation~(RPA)~\citep{ivanvcic2019robotic,wewerka2020robotic,agostinelli2020towards,ferreira2020evaluation}), the \textit{de facto} predominant automation technology, thus has been widely applied, which automates a process by orchestrating several software by manual-crafted rules into a solidified workflow for efficient execution~\citep{zapier,n8n,unipath}.
Despite its strides, \textbf{robotic process automation merely offloads simple and mechanical human labor, while processes requiring human intelligence still necessitate human labor.}
First, as Figure~\ref{fig:framework-comparison} shows, while RPA workflows can perform processes automatically, their construction still requires human intelligence for elaborate design.
Second, many tasks performed by humans are characterized by their flexible and complex nature while workflows are limited to mechanistically replicating human behavioral processes, posing challenges in automating intricate processes that demand dynamic decision-making capabilities during execution.

%

With the rapid development of Large Language Models~(LLMs)~\citep{openaichatgptblog,openai2023gpt4}, LLMs are emerging with intelligence that was previously exclusive to human beings~\citep{wei2022emergent}.
Recently, LLM-based agents have garnered significant attention from the research community~\citep{xi2023rise,wang2023survey,yao2022react,shinn2023reflexion,sumers2023cognitive,qin2023toolllm,ye2023large}. 
LLM-based agents have demonstrated a certain level of human intelligence, being capable of using tools~\citep{schick2023toolformer,qin2023tool,qin2023toolllm}, creating tools~\citep{qian2023creator,cai2023large}, playing games~\citep{wang2023voyager,chen2023agentverse}, browsing website~\citep{nakano2021webgpt,qin2023webcpm,yao2022webshop}, developping software~\citep{qian2023communicative} akin to humans.
%
%
Consequently, a meaningful inquiry naturally emerges: \textbf{Can LLM-based agents advance automation in processes necessitating human intelligence, further liberating human beings?}


In this paper, we propose \agenticauto~(APA), a novel process automation paradigm that overcomes the two aforementioned limitations of automation. 
%
%
(1) Agentic Workflow Construction: Upon receiving human requirements or instructions, LLM-based agents elaborately construct the corresponding workflows instead of humans. 
If a process involves dynamic decision-making, agents should recognize which part of this process needs the dynamic decision-making and then orchestrate agents into the workflow.
%
%
(2) Agentic Workflow Execution: Workflows should be monitored by agents and once the workflow is executed in the dynamic part, agents would intervene to handle the dynamic decision-making.
%


\begin{table*}[!tb]
    \centering
    \begin{minipage}{0.75\linewidth}
        \resizebox{0.9\linewidth}{!}{
            \begin{tabular}{lcccc}
                \toprule
                \multirow{2}{*}{\textbf{Paradigm}} & \multicolumn{2}{c}{\textbf{Efficiency}} & \multicolumn{2}{c}{\textbf{Intelligence}} \\
                \cmidrule{2-5}
                & Data Flow & Control Flow & Data Flow & Control Flow \\
                \midrule
                  RPA & \cmark & \cmark & \xmark & \xmark \\
                  LLM-based Agents & \xmark & \xmark & \cmark & \cmark  \\ 
                  \hdashline
                  APA & \cmark & \cmark & \cmark & \cmark \\
                  \quad \dataagent & \cmark & \cmark & \cmark & \xmark \\
                  \quad \controlagent & \cmark & \cmark & \xmark & \cmark \\
                 \bottomrule
            \end{tabular}
        }
    \end{minipage}%
    \hfill
    \begin{minipage}{0.25\linewidth}
        \centering
        \includegraphics[width=0.8\linewidth]{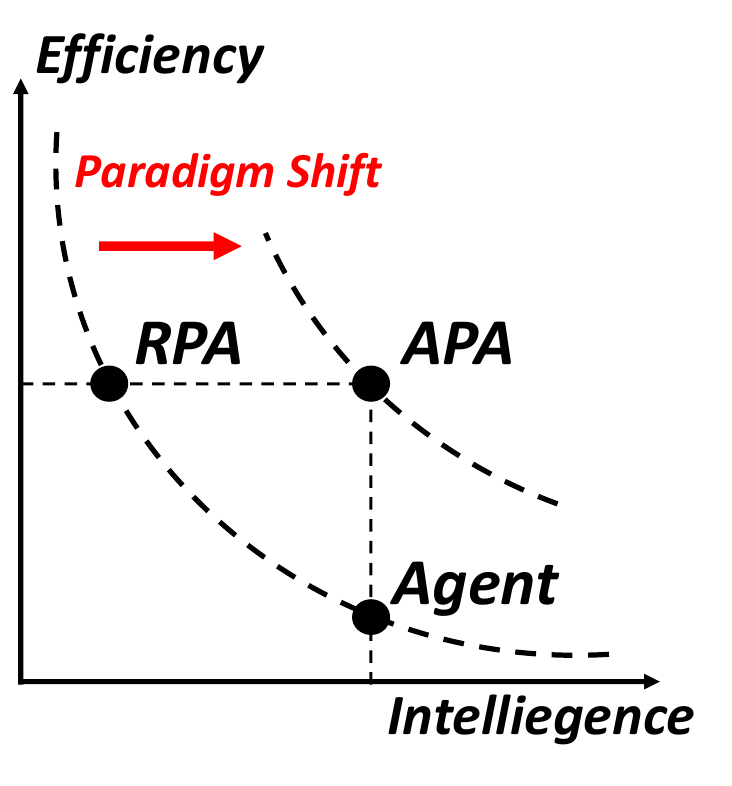}
    \end{minipage}%
    \caption{A comparison between robotic process automation and agentic process automation in terms of efficiency and flexibility.}
    \label{tab:framework_comparison}
\end{table*}

To explore the feasibility of APA, we instantiate \model, an LLM-based agent that integrates the agentic workflow construction and agentic workflow execution in a unified framework.
%
%
For agentic workflow construction, to make LLM-based agents understand and generate workflows, we design \textbf{\workflow} based on the JSON structure and Python code, stemming from the realization that LLMs are pretrained on coding corpus.
Specifically, it adopts JSON structure to organize the input and output data for each software for data standardization and uses Python code to implement process control logic to orchestrate software~(see in Figure~\ref{fig:workflow-language}).
%
%
Upon receiving a specific task, \model is able to generate the corresponding workflow language to facilitate the construction of the requisite workflow.
For agentic workflow execution, dynamic decision-making in workflows encompasses two aspects: 
%
%
(1) Data flow: complex data processing (e.g., writing data analysis reports) often exceed the capacity of rule-based systems and thus agents must intervene to effectively manage these intricate processes.
(2) Control flow: complex tasks may involve intricate conditional branches and loops, which surpass the expression ability of rules. 
In such cases, agents need to function as controllers to dynamically determine the subsequent actions.
Hence, we design two types of dynamic decision-making agents: \textbf{\dataagent} acts as a data processing to handle intricate data processes dynamically and \textbf{\controlagent} functions as a condition expression that enables the dynamic determination of subsequent branches for execution.
%
%
Confronted with complex tasks that need intelligence, \model can orchestrate these two agents into the workflows during construction and handle complex circumstances purposefully during execution, offloading the intelligent labor~(see in Table~\ref{tab:framework_comparison}).

To empirically validate our approach, we conduct proof-of-concept experiments to showcase that \model is able to construct workflows based on human instructions and handle the dynamic decision-making part of the process by utilizing agents in the workflow. 
We further discuss the relationship between \model with existing research areas, including Tool Learning~\citep{qin2023tool,qin2023toolllm}, Process Mining~\citep{tiwari2008review,van2012process,turner2012process}, Safety~\citep{cummings2004automation} and etc.
Our contributions are listed as follows:
\begin{itemize}[topsep=1pt, partopsep=1pt, leftmargin=12pt, itemsep=-1pt]
    \item We propose \agenticauto, a new process automation paradigm that integrates LLM-based agents to further offload the intelligent labor of humans.
    \item We instantiate \model, in which \workflow is desgined for LLM-based agents to construct workflows and \dataagent and \controlagent are orchestrated into workflows to handle the dynamic decision-making process part purposefully.
    \item We demonstrate of feasibility of our \model through proof-of-concept case analyses and the exploration of potential and opportunities of \agenticauto across various research domains including tool learning, process mining, safety, etc.
\end{itemize}

\section{Methodology}




Workflow is widely-used in RPA to solidify the process by a software invocation graph, where nodes represent a software operation and edges signify topology of the process of execution.
To achieve the solidification, a data flow and a control flow are involved to within the workflow.
Data flow describes how data is passed and processed within a series of software and control flow describes the order of software to execute.
In this section, we first introduce \workflow to express the data flow and control flow, and then we further detail how to integrate agents into workflows to bring flexibility into workflows.
Finally, we detail the workflow construction and execution procedure about how \model works.

\subsection{\workflow}

\begin{figure}[!t]
\centering
\includegraphics[width=1.0\linewidth]{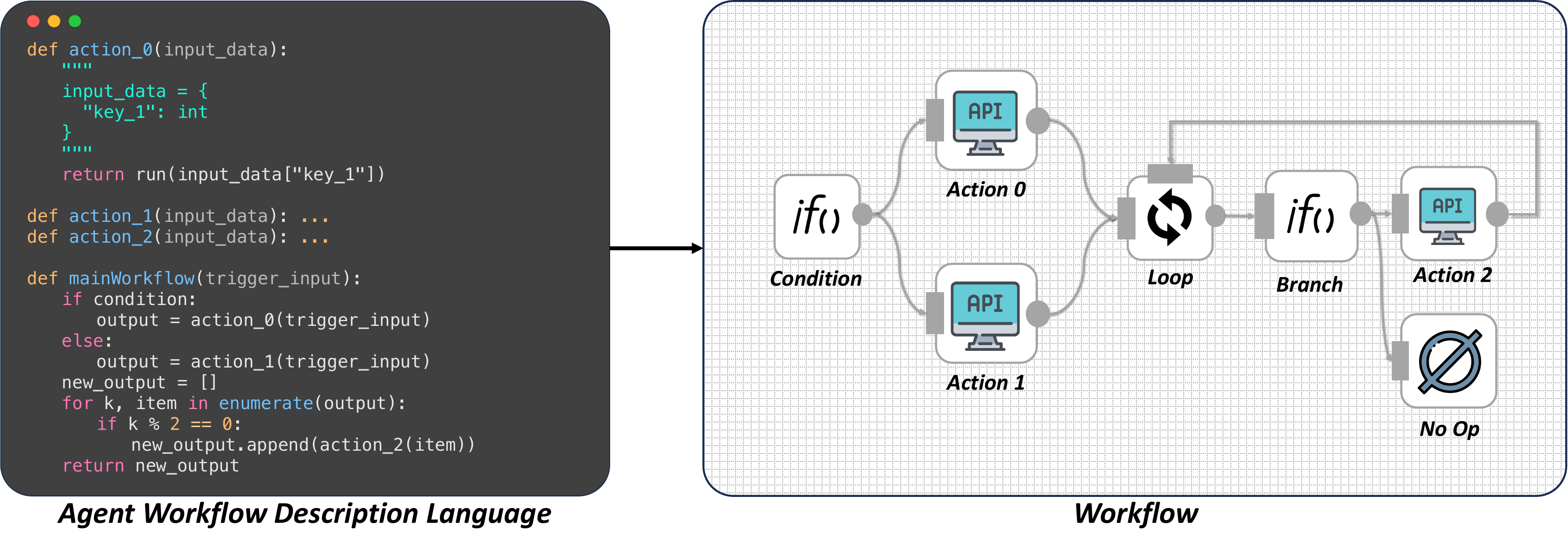}
\caption{Illustration of \workflow.}
\label{fig:workflow-language}
\end{figure}

%

As workflow is a graph-based representation approach for RPA to solidify the process, it is inadaptive to LLMs to understand and generate workflows.
Thus, we we elaborately design \workflow for LLM-agents to conveniently solidify workflows based on the characteristics of coding pretraining.
Specifically, we adopt JSON structure to describe data flow and Python code to describe control flow.
Figure~\ref{fig:workflow-language} gives the illustration of \workflow.





\paragraph{JSON Structure for Data Flow}

To solidify a workflow, the data format through software should be standardized to ensure the automatic data process, free from unnecessary agent interventions.
We adapt the JSON structure to organize the input/output data of all actions in the workflow.
As Figure~\ref{fig:workflow-language} shows, the input data is formatted in a key-value-paired dictionary.
Every data should be assigned a specific key, making it easy to parse and manipulate.
When transferring data between different software, the JSON structure is convenient to index the specific data field.
%
%
%
Only when the input and output of all software are strictly standardized, promoting consistency across different software of the workflow, thereby reducing the likelihood of data interpretation errors or discrepancies.
%

\paragraph{Python Code for Control Flow}

For complex tasks, the corresponding workflows usually involve complex control logic, including conditional branches, loops, or sub-workflow execution.
Conventional RPA methods commonly design graph-based representations for human developers to describe the control flow~\citep{zapier,n8n,unipath} but its expression ability for complex workflow is limited and it is also not suitable for LLM-based agents to understand and generate.
As Python programming language supports complex control logic and more importantly and it is learned by LLMs during the pre-training phase, we use Python to describe the control flow.
As a high-level programming language, Python offers a rich set of primitives and features, providing greater expressive capability to describe complex control logic.
A workflow is composed of a Python file, with each software operation aligned to a Python function called \textit{action}.
The corresponding input/output data is mapped into the parameters and return values of the function.
Thus, a series of actions (i.e., software) are described as sequential function callings in Python.
The if-else statement and for/while statement in Python can be used to implement complex logic control flow.
Finally, the workflow is encapsulated within a main Python function (i.e., \texttt{mainWorkflow}).
Furthermore, as Python supports the nested function calling, different workflows can also be composed together by calling workflow function to construct a complex workflow.
During workflow execution, we utilize a Python executor, starting from the main workflow function (\texttt{mainWorkflow}) as the entry point and execute each functions sequentially, ultimately completing the entire workflow execution.

\subsection{Agent-Integrated Workflow}

\begin{figure}[!t]
\centering
\includegraphics[width=1.0\linewidth]{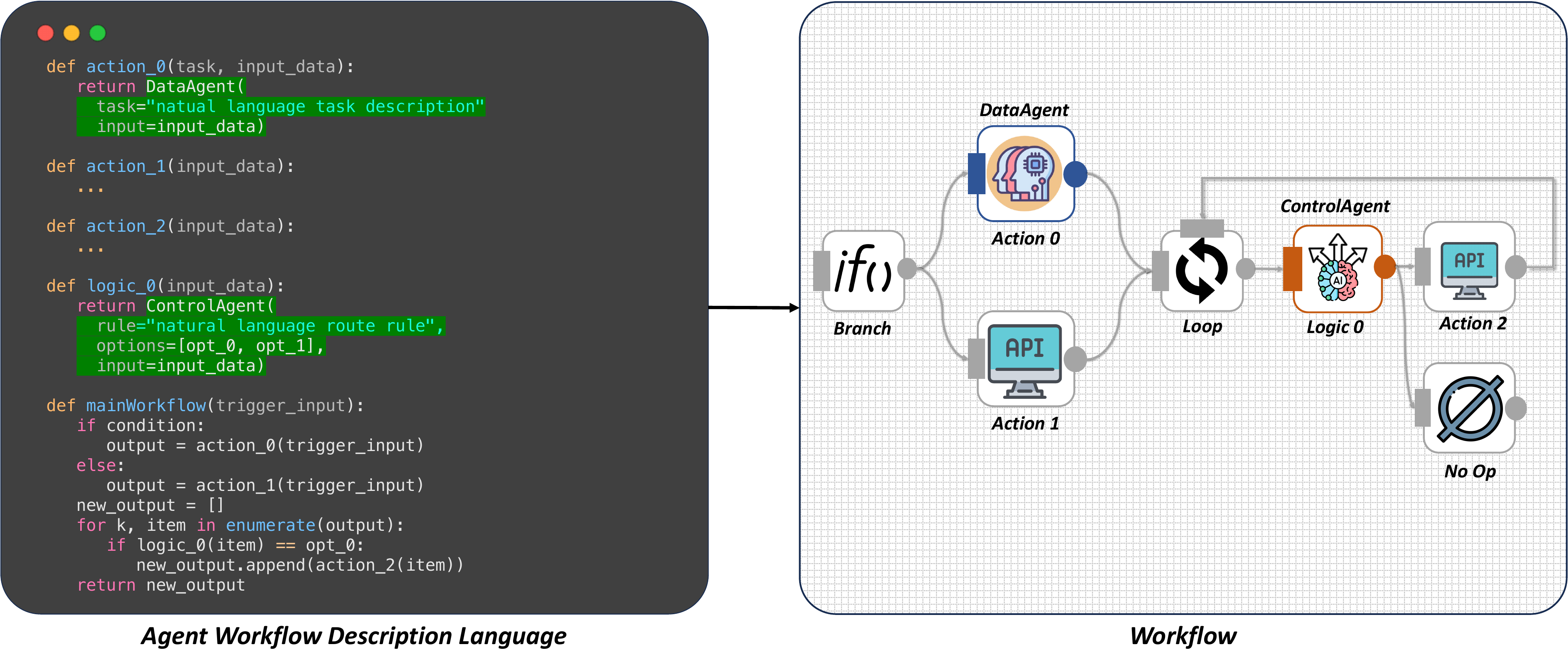}
\caption{Illustration of \workflow with \dataagent and \controlagent.}
\label{fig:agent-integrated-workflow}
\end{figure}

%
%
%
%
As many real-world tasks with flexibility and complexity nature involve dynamic decision-making process, we devise \dataagent and \controlagent which can be orchestrated into workflows to handle the dynamic part during execution.
Figure~\ref{fig:agent-integrated-workflow} gives the illustration.




\paragraph{\dataagent}

To achieve complex data process, we devise \dataagent, which acts as an action that is operated by an LLM-based agent.
As Figure~\ref{fig:agent-integrated-workflow} shows, it supports inputting a task description based on natural language and then accomplishing this task autonomously based on the intelligence of the agent. 
During execution, this function initiates a ReACT-based agent~\citep{yao2022react} to fulfill the task. 
\begin{equation}
    \mathtt{output} \leftarrow \verb|DataAgent|(\mathtt{task}, \mathtt{input})
\end{equation}
Although the function is actually operated by agents, its input/output data are still organized by JSON to make it can be orchestrated into existing workflows to connect with other actions.
By incorporating the \dataagent, the workflow provides support for enhanced flexibility for data flow, enabling the handling of intricate data processing demands.

\paragraph{\controlagent}

In addition to serving as the action, agents can be further involved in the control flow to schedule the execution logic. 
We introduce \controlagent into the control flow, allowing it to substitute a selection expression. 
As Figure~\ref{fig:agent-integrated-workflow} shows, \controlagent contains a pre-generated judgment criterion based on natural language and several execution branch candidates.
\begin{equation}
    \mathtt{opt} \leftarrow \verb|ControlAgent|(\mathtt{task}, \mathtt{input}, [\mathtt{opt_1}, \mathtt{opt_2}, \cdots, \mathtt{opt_n}])
\end{equation}
During execution, the agent can make a decision based on the input data to decide which branch will be executed subsequently, influencing the control flow of the workflow.

\subsection{Workflow Construction}

\begin{figure}[!t]
    \centering
    \includegraphics[width=1.0\linewidth]{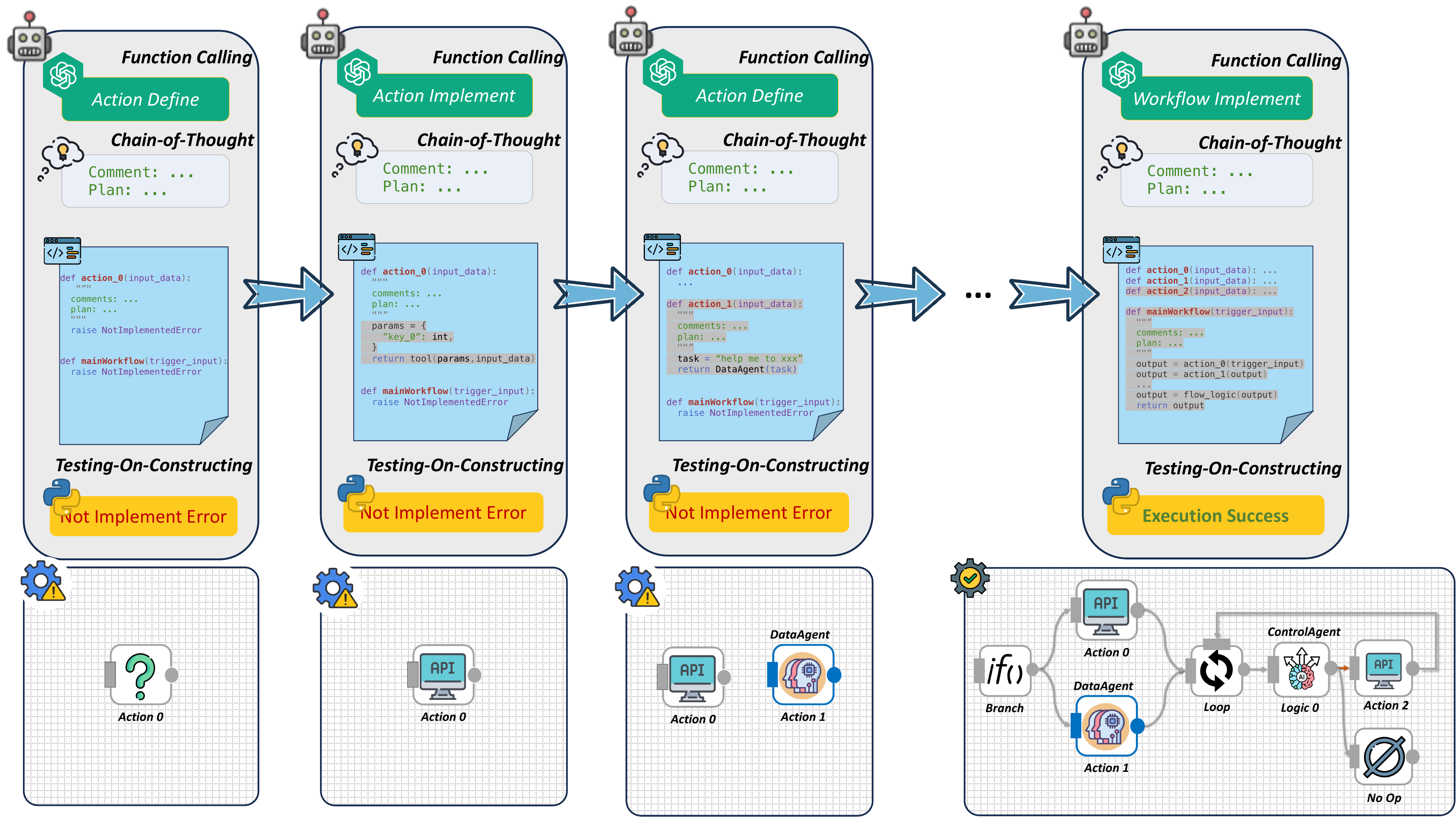}
    \caption{The Illustration of the workflow construction procedure of \model.}
    \label{fig:workflow-construction}
\end{figure}

As the workflow is represented as JSON structure and Python code, the workflow construction is formulated as a code generation task. 
%
%
As Figure~\ref{fig:workflow-construction} demonstrates, the workflow construction procedure contains four iterative operations:
\begin{itemize}[topsep=1pt, partopsep=1pt, leftmargin=12pt, itemsep=-1pt]
    \item \texttt{action\_define}: It determines which action is selected to add into the workflow.
    \item \texttt{action\_implement}: It first transforms the action into the Python function by determining its input/output data format in JSON structure and then implement the data process program in Python code.
    \item \texttt{workflow\_implement}: As workflows are represented as \texttt{mainWorkflow} functions, this operation refers to providing an implementation for it to orchestrate the entire workflow.
    \item \texttt{task\_submit}: It is used to denote the termination of the workflow construction.
\end{itemize}

In practice, we employ \texttt{OpenAI GPT-4} as the backbone of \model to generate the workflow language and further incorporated several techniques to enhance the workflow generation capabilities:
\begin{itemize}[topsep=1pt, partopsep=1pt, leftmargin=12pt, itemsep=-1pt]
    \item {Testing-on-Constructing}: During the construction, \model tends to test each function or entire workflow, which ensures the validation of the constructed workflow before execution.
    \item {Function Calling}: The aforementioned four operations are defined as \texttt{function} in \texttt{GPT-4} to use \texttt{Function Calling} to explicitly control the whole construction procedure, benefiting controllable generation.
    \item {Chain-of-Thought}: When implementing each function, \model requires to provide a comment (explaining the purpose of this function) and a plan (indicating what the subsequent operations should be done next), which aids in enhancing the workflow code generation performance.
    
\end{itemize}

\subsection{Workflow Execution}

The workflow execution procedure is based on Python interpreter.
Given a workflow language, once this workflow is triggered, its corresponding \texttt{mainWorkflow} function is selected as the entry point to begin the execution procedure.
The execution procedure follows the Python code execution rule, i.e., executing according to the line order sequentially.
%
%
Once the \texttt{mainWorkflow} function returns, the workflow execution is finished successfully.

\section{Proof-of-Concept Experiment}

To validate the feasibility of \agenticauto, we conduct proof-of-concept experiment based on n8n~\footnote{https://n8n.io}, an open-source workflow platform.
Each APP (i.g., software) in the n8n platform is encapsulated as an action in the workflow and thus the core of the workflow construction is to orchestrate these APPs to achieve certain tasks.
We implement our proposed \model based on \texttt{GPT-4}.
We construct a case about the commercial scenario to explain how our \model works in detail.

\subsection{Task Construction}
\begin{figure}[!h]
\centering


\includegraphics[width=1.0\linewidth]{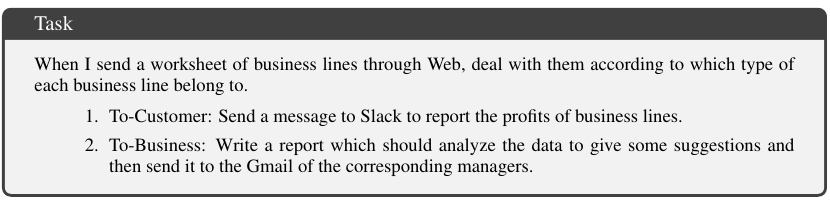}
\end{figure}
We present a typical commercial scenario where a business department manager seeks to extract diverse business line data from Google Sheets. 
Upon identifying a business line is To-Customer, a message is dispatched to the Slack channel. 
\begin{figure}[!th]
    \centering
    \includegraphics[width=0.8\linewidth]{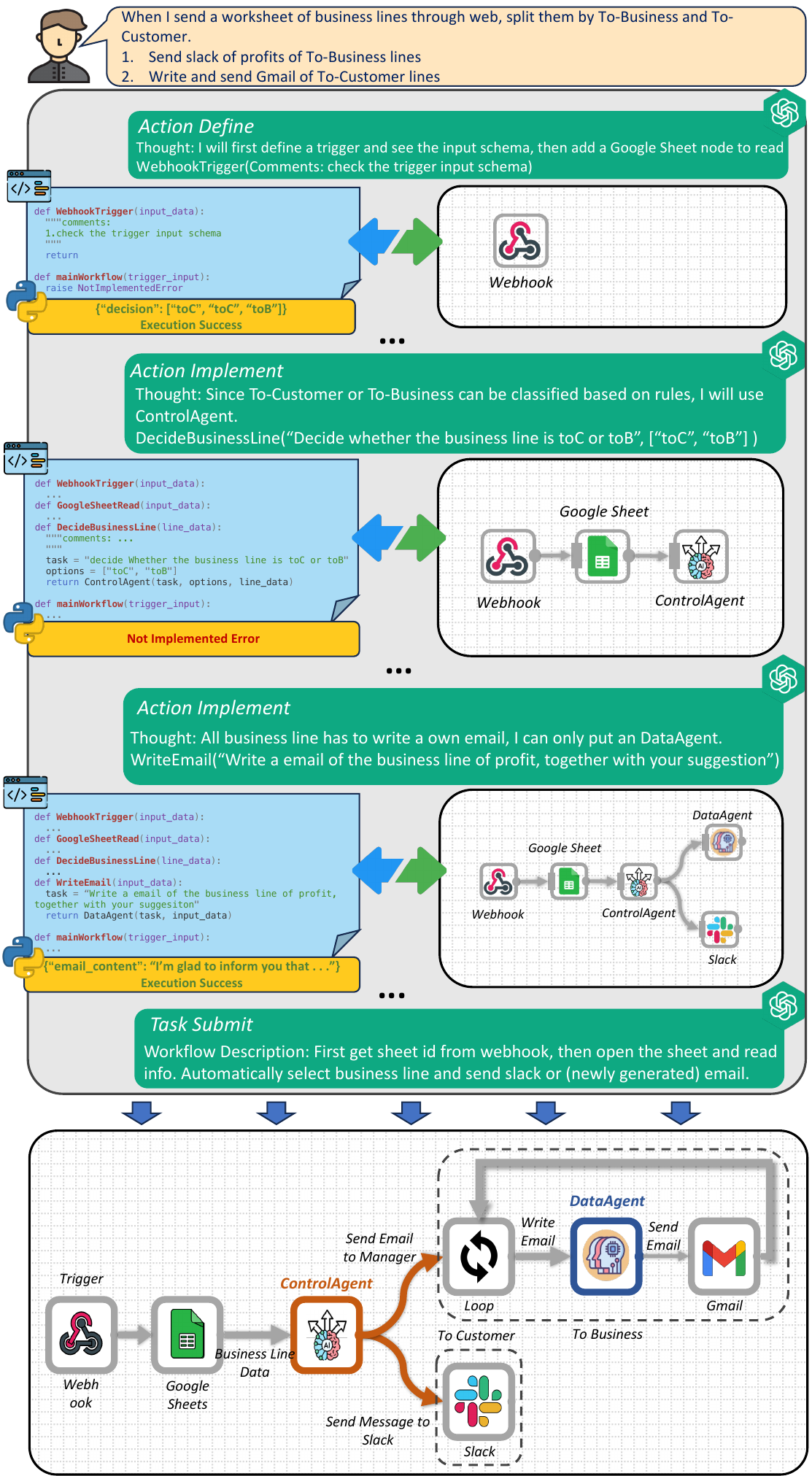}
    \caption{The Illustration of the workflow construction procedure of \model for case analysis.}
    \label{fig:construction-case}
\end{figure}
In the case of a business line as To-Business, an email is sent to the respective manager, encompassing an assessment of the business line and a concise overview of its profitability.
This exemplifies the capabilities achievable through the integration of workflow and agent:
\begin{itemize}[topsep=1pt, partopsep=1pt, leftmargin=12pt, itemsep=-1pt]
    \item Given the similar format of the content in Google Sheets, this task exhibits a high level of reusability, applicable to various business departments with analogous requirements.
    \item Distinguishing whether a business line pertains To-Customer or To-Business based solely on its description is a complex task not easily handled through rule-based judgments. Agent integration becomes imperative for dynamic branching.
    \item Generating corresponding emails for the relevant business lines is a dynamic requirement necessitating flexible agent-based data processing.
\end{itemize}

\subsection{Workflow Construction}
Figure~\ref{fig:construction-case} gives the visualization of the constructed workflow by \model.
\model constructs a workflow with seven nodes for this case, including a \dataagent node and a \controlagent node. 
As the user says that he will send data through Web, \model decide to define an action named \texttt{WebhookTrigger} as the trigger to for the workflow.
Then it implement \texttt{GoogleSheetRead} action to read data from Google Sheets according to the user description.
Since it should execute different actions according to whether the business line belong to To-Business or To-Customer which needs to understand the meaning of each business line, \model define a \controlagent which aims to \textit{decide whether the business line is toC or toB} to judge what next action to execute.
If the business line belong to To-Customer, as the user description, \model implements the Slack action which send the corresponding profits into Slack.
If the business line belong to To-Business, it needs to write a detailed report to analyze the specific data in Google Sheets and give some suggestions.
Thus, \model implement a \dataagent \texttt{WriteEmail} which task is to \textit{Write a email of the business line of profit, together with your suggestions}.
Then, a Gamil APP is following implemented to send the generated email to the corresponding managers.
As there may exist multiple To-Business data in Google Sheets, \model further add a Loop in workflow to deal with these data iteratively.
Finally, as the workflow is constructed completely, \texttt{task\_submit} is operated by \model to end the construction procedure.

%
%
%
%
%
%
%

\subsection{Workflow Execution}

\begin{figure}[!t]
    \centering
    \includegraphics[width=1.0\linewidth]{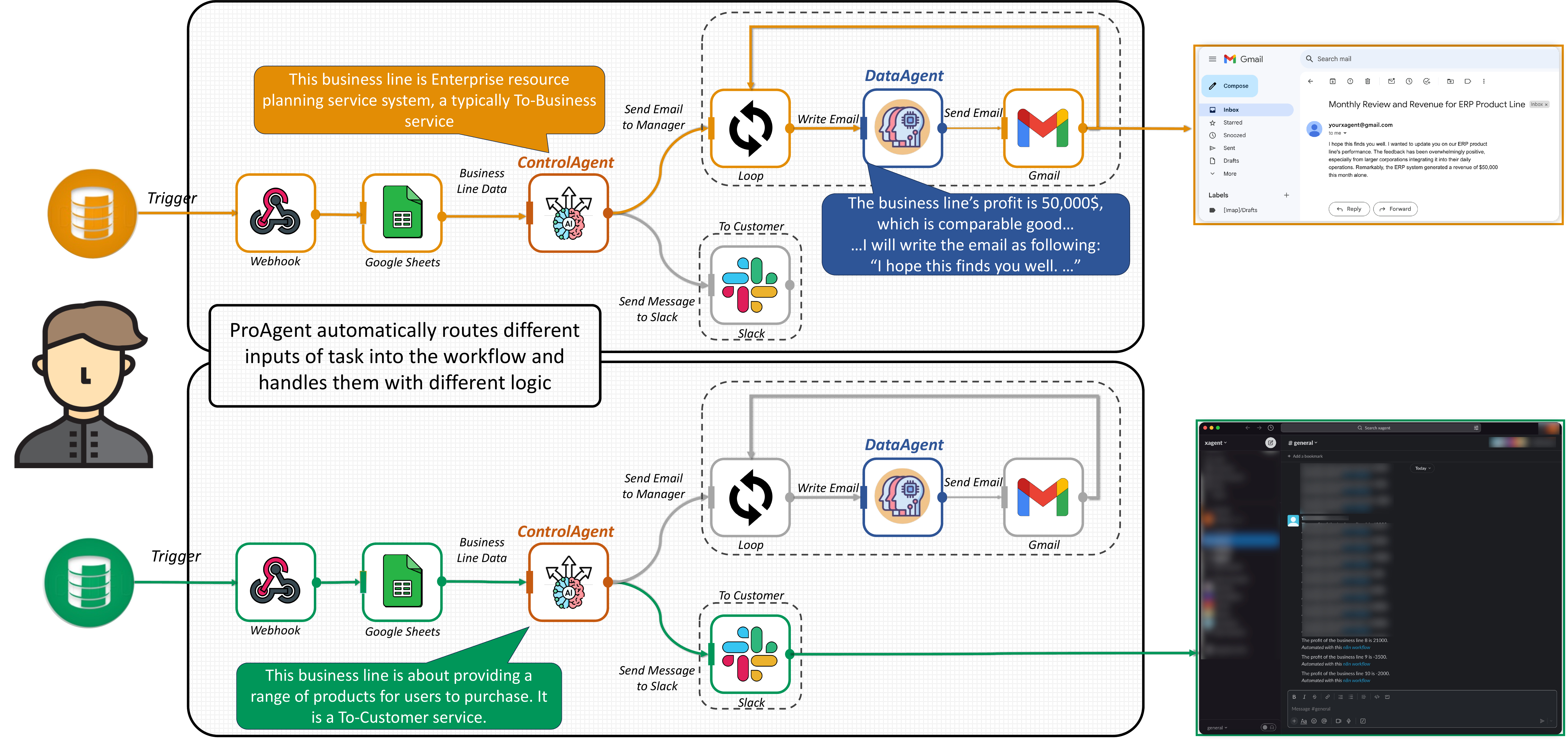}
    \caption{The Illustration of the workflow execution procedure of \model for case analysis.}
    \label{fig:execution-case}
\end{figure}

Figure~\ref{fig:execution-case} illustrates two execution cases for the constructed workflow.
These two cases demonstrate a To-Customer and a To-Business line respectively. 
It is obviously shown that the \controlagent successfully distinguish which type of two business lines belong to.
For the first one, the description of this business line is:
%
%
\begin{figure}[!h]
\centering
\includegraphics[width=1.0\linewidth]{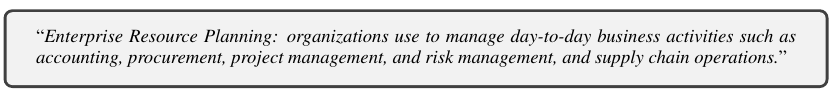}
\end{figure}

%
\controlagent distinguish that \textit{This business line is Enterprise resource planning system, a typically to-business service}.
After that, the \dataagent is executed to write an email to send: 
%
%
\begin{figure}[!h]
\centering
\includegraphics[width=1.0\linewidth]{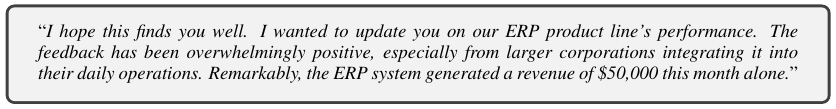}
\end{figure}

For the second one, its description is:
%
%
\begin{figure}[!h]
\centering
\includegraphics[width=1.0\linewidth]{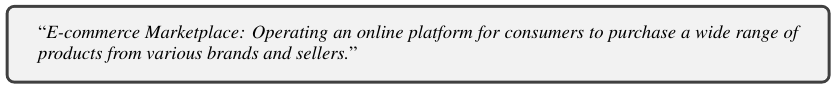}
\end{figure}

\model extracts some key words such as ``products'' and ``purchase'' to decide that this business line belongs to To-Customer type.
Then, a simple message ``\textit{The profit of the business line 9 is -3500\$.}'' to the Slack.

\section{Discussion}

In this section, we discuss the relationship between APA and \model technologies and the field of tool learning, process mining, etc, along with potential research directions. 
Subsequently, we discuss the issue of automation bias that may arise from APA technologies and delve into the unique value humans should bring in the era of automation.

\subsection{ProAgent as ToolAgent: Integrating Tool Utilization and Tool Creation}
Existing work has shown that LLM-based agents own powerful capability to utilize external tools (e.g., search engine, web browser, calculator, python interpreter, etc)~\citep{ivanvcic2019robotic,wewerka2020robotic,agostinelli2020towards,ferreira2020evaluation}.
By integrating the external tools, agents can extend its capability from text generation to manipulate tools to impact environments and even the real world.
In addition to their adeptness in utilizing existing tools, recent research has showcased their capability to create novel tools for specific tasks~\citep{qian2023creator,cai2023large}. 
In a workflow, a series of operations are involved, and when these operations involve invoking external tools, the execution process of this workflow can be viewed as the process of tool utilization~\citep{nakano2021webgpt,yao2022webshop,schick2023toolformer,qin2023tool,qin2023toolllm,ye2023large}.
In the context of workflow management, the execution of a series of operations through the invocation of external tools can be perceived as a process of tool utilization. 
Orchestrating multiple operations (i.e., tools) into a coherent workflow to address specific tasks can be viewed as the process of tool creation, i.e., combining multiple tools to form a new tool.
In this context, \model can be understood as an autonomous process that seamlessly integrates tool creation and utilization.

\subsection{Process Mining: Discover, Analyze, and Improve Workflows}
In our \model, the workflow construction process is primarily driven by the agent solely, potentially yielding sub-optimal workflows. 
Integrating the principles of Process Mining~(PM)~\citep{tiwari2008review,van2012process,turner2012process} into this framework can facilitate the discovery of valuable workflows from historical records. 
Process Mining is a data-driven technique used to discover, analyze, and improve existing workflows. 
It involves the extraction of insights and knowledge from historical records generated during the execution of various workflows. 
This method enables to gain a comprehensive understanding of workflows constructed by agents, identify inefficiencies, bottlenecks, and deviations from the intended workflow, and subsequently optimize their operations for better performance and efficiency. 
Moreover, optimized workflows can serve as effective training data to enhance the agent's workflow construction capabilities.

\subsection{Ethical and Safety Concerns: Agent aggravates Automation Bias}
The advent of agent technology offers the potential for models to tackle complex tasks. 
However, on the flip side, it can also lead to an over-reliance on agents by humans. 
People might opt to trust agent decisions even when these decisions conflict with their own views, which is known as Automation Bias~\citep{cummings2004automation} and has been observed in domains such as Clinical decision support systems~\citep{goddard2012automation}.
Under LLM-based agent technologies, this problem becomes even more pronounced. 
Humans may shift their trust in the stability of traditional rule-based workflows to agents, mistakenly believing that the agent's decision-making processes are equally reliable, especially hallucination~\citep{maynez2020faithfulness,zhang2023siren,ji2023survey}.
%
%
Future research is necessary to prioritize the development of safer, more trustful, more interpretable agentic process automation.



\subsection{Human Advantage: Rethinking the meaning of human labor}
APA introduces the intelligence of elaborate design in workflow construction and dynamic decision-making in workflow execution into process automation, which can offload the heavy human labor in RPA.
Now, the more pertinent question is: ``\textit{What tasks should remain human-driven?}''
There are processes that inherently benefit from human intuition, experience, and creativity. 
For these tasks, humans play a crucial role that can't be easily supplanted by machines. 
While automation might offer efficiency, it can't replicate the nuanced understanding and innovative solutions that a human brings to the table.
The paradox of human involvement, where human intervention can improve outcomes, stands in contrast to the earlier mentioned pitfalls of automation bias.

The next frontier in APA involves discerning which processes can be wholly automated and which require human oversight or intervention. 
We must remember that the ultimate goal of automation is to amplify productivity, not to supplant humans entirely. 
The challenge lies in facilitating a symbiotic relationship between humans and machines, where neither is completely excluded in favor of the other. 
Drawing from the perspective of Steve Jobs, the future should see humans focusing on what they do best: applying their unique intelligence and creativity where it matters most. 
APA demands a recalibration, where automation serves humanity, and humans, in turn, elevate the capabilities of automation.

\section{Related Work}

\paragraph{Robotic Process Automation}

Robotic process automation~(RPA)~\citep{ivanvcic2019robotic,hofmann2020robotic,tiwari2008review,scheer2004business}, as the fashion automation paradigm, primarily employs software robots to either automate access to software APIs or simulate user GUI interactions to accomplish tasks through multiple software.
Unlike traditional automation techniques, RPA emulates the way humans use software, directly tapping into existing software assets without the need for transformation or additional investment. 
Thus, RPA has gained substantial attention in recent years as an effective technology for automating repetitive and rule-based tasks typically performed by human workers~\citep{zapier,n8n,unipath}. 
RPA are primarily designed to automate repetitive tasks using predefined rules and workflow templates, which needs heavy human labor to design and implement workflows.
Still, due to the workflows are driven by manual-crafted rules, it struggles to handle those complex tasks that needs dynamic decision-making.

Recently, there has been a growing interest in integrating RPA with AI technique, leading to various terminologies and definitions. 
For instance, \textit{Intelligent Process Automation}~(IPA)~\citep{ferreira2020evaluation, chakraborti2020robotic} and \textit{Cognitive Automation}~(or RPA 4.0)~\citep{lacity2018robotic}, aim to amalgamate AI techniques in the phases of RPA, e.g., data format transformation~\citep{leno2020automated}, workflow optimization~\citep{chakraborti2020d3ba}, conversational assistant~\citep{moiseeva2020multipurpose}, demonstration-to-process translation~\citep{li2019interactive}, etc.
Nevertheless, these work still utilizes traditional deep learning technique (e.g., RNN~\citep{han2020automatic}) or even machine learning technique (e.g., Monte Carlo Tree Search~\citep{chen2020monte}) into RPA.
More importantly, they just utilizes AI technique into some specific fragments of RPA (e.g., data format transformation~\citep{leno2020automated}).
In contrast, our work \agenticauto takes the lead to integrate the most intelligent AI model, large language models, into RPA.
As a result, it is the inaugural exploration into agentic techniques in both the autonomous generation of workflows and Agent-driven workflow execution to endow them with intelligence.

\paragraph{LLM-based Agents}

Large language models~(LLMs), as significant milestones of artificial intelligence, unveil the remarkable capability on a wide range of tasks~\citep{openaichatgptblog,openai2023gpt4}.
Recently, LLM-based agents emerge to extend LLMs with external tools to interact with the environment to achieve real-world tasks.
Early research work attempt to prompt LLMs to generate the action according to the observation of environment~\citep{nakano2021webgpt,huang2022language,ahn2022can,schick2023toolformer,qian2023communicative,chen2023agentverse}.
Such a manner tends to struggle when facing intricate tasks that need long-term planning and decision-making.
To address this issue, ReAct~\citep{yao2022react} proposed a dynamic task-solving approach that makes agents generate thought for each action to form a reasoning chain, enabling flexible reasoning-guided, trackable, and adjustable actions, resulting in notable improvements compared to act-only methodologies.
Based on the dynamic task-solving manner, many agents are proposed subsequently to improve agent capability in different aspects, e.g., reflection~\citep{shinn2023reflexion}, planning~\citep{yao2023tree,hao2023reasoning,besta2023graph,sel2023algorithm}, tool learning~\citep{schick2023toolformer, patil2023gorilla,qin2023tool,qin2023toolllm,qian2023creator}, multi-agents~\citep{park2023generative,qian2023communicative}, etc.
However, all the existing ReACT-based agent methods are restricted to linearly generate decision-making, resulting in lower operational efficiency.
In this paper, we propose \model that explores to enhance the efficiency of the dynamic task-solving approach by recognizing which part of the workflow needs the intelligence involves and integrating agents to handle these parts purposefully.
%

\section{Conclusion}


In this research, we present a novel process automation paradigm, \agenticauto, to address the limitations of robotic process automation technologies in handling tasks requiring human intelligence by harnessing the capabilities of LLM-based agents to integrate them into the workflow construction and execution process.
Through the instantiation of \model, we illustrated how LLM-based agents can feasibly manage complex decision-making processes, thereby offloading the burden of intelligent labor from humans.
Our proof-of-concept experiment provided evidence of the feasibility of \agenticauto in achieving efficiency and flexibility in process automation. 
Our findings contribute to the growing body of research in the field of intelligent automation and underscore the significant role that LLM-based agents can play in enhancing the efficiency and flexibility of various industries. 
%
%
As the adoption of automation technologies continues to expand, we anticipate that the APA framework can serve as a catalyst for further advancements in the automation landscape, leading to increased efficiency, reduced human intervention, and ultimately, a more streamlined and intelligent workflow ecosystem.

\bibliography{ref}
\bibliographystyle{iclr2024_conference}

\appendix

\end{document}